\documentclass[runningheads]{llncs}
\usepackage[T1]{fontenc}
\usepackage{graphicx}
\usepackage[most]{tcolorbox}
\usepackage[inline]{enumitem}
\usepackage{svg}
\usepackage{url}
\usepackage{hyperref}
\usepackage{algorithm}
\usepackage[noend]{algpseudocode}
\usepackage[title]{appendix}
\usepackage[caption=false]{subfig}
\usepackage[inline]{enumitem}
\usepackage{wrapfig}

\newtcolorbox[blend into=figures]{myfigure}[2][]{size=fbox,float=htb,capture=minipage,
title={#2},every float=\centering,#1}

\begin{document}
\title{TIC: Translate-Infer-Compile for accurate “text to plan” using LLMs and Logical Representations}
\titlerunning{Translate-Infer-Compile}
\author{Sudhir Agarwal \and Anu Sreepathy}
\institute{Intuit AI Research, Mountain View, CA, USA \\
\email{\{sudhir\_agarwal, anu\_sreepathy\}@intuit.com}
}
\authorrunning{S. Agarwal et al.}
\maketitle              % typeset the header of the contribution
\begin{abstract}
We study the problem of generating plans for given natural language planning task requests. On one hand, LLMs excel at natural language processing but do not perform well on planning. On the other hand, classical planning tools excel at planning tasks but require input in a structured language such as the Planning Domain Definition Language (PDDL). We leverage the strengths of both the techniques by using an LLM for generating the PDDL representation (task PDDL) of planning task requests followed by using a classical planner for computing a plan. Unlike previous approaches that use LLMs for generating task PDDLs directly, our approach comprises of (a) \textbf{translate:} using an LLM only for generating a logically interpretable intermediate representation of natural language task description, (b) \textbf{infer:} deriving additional logically dependent information from the intermediate representation using a logic reasoner (currently, Answer Set Programming solver), and (c) \textbf{compile:} generating the target task PDDL from the base and inferred information. We observe that using an LLM to only output the intermediate representation significantly reduces LLM errors. Consequently, TIC approach achieves, for at least one LLM, high accuracy on task PDDL generation for all seven domains of our evaluation dataset.

\end{abstract}
\section{Introduction}
\label{sec:introduction}
\begin{figure}[t]
  \centering
  \includegraphics[width=\linewidth]{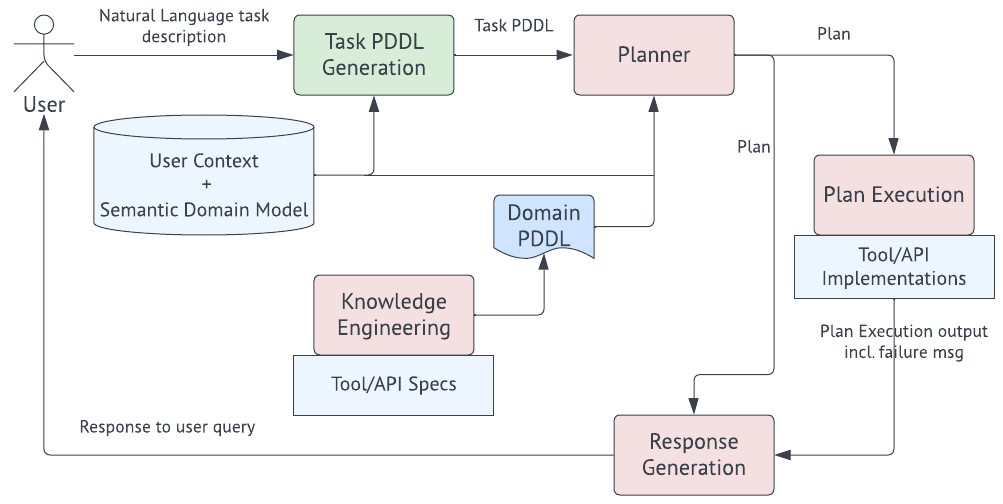}
  \vspace{-.5cm}
  \caption{An overview of planning-based response generation}
  \vspace{-.5cm}
  \label{fig:e2e-Diagram}
\end{figure}
Customers of large organizations have a variety of questions or requests (collectively known as queries in the following) pertaining to the organization's domain of operation. %Providing accurate responses to such user queries requires a thorough analysis of the user's context, product features, domain knowledge, and organizational policies.%
Certain user queries such as how-to questions and state changing requests require dynamic composition of possible actions (i.e., planning). Recently, transformer-based large language models (LLMs) have shown wide success on many natural language understanding and translation tasks, also demonstrating some general reasoning and planning capability on diverse tasks without having to be retrained~\cite{ahn2022i,huang2022inner,zeng2022socratic}. However, LLMs are known to perform only shallow reasoning and cannot find complex plans~\cite{valmeekam2023large,valmeekam2023planning,openai2023gpt4,silver2022pddl}. Some approaches such as~\cite{DBLP:conf/nips/GuanVSK23,zhou2023isrllm} introduce frameworks for corrective feedback and iterative self-refinement of LLM generated plans. Other frameworks such as LangChain~\cite{langchain} and Gorilla~\cite{patil2023gorilla} enable developers to combine LLMs with external tools or APIs. Such frameworks depend on LLMs for selecting and composing tools and either do not scale well beyond a small set of tools or require fine-tuning of LLMs.
On the other hand, classical planners can scale better in the number of actions they support without compromising on the accuracy~\cite{DBLP:conf/ijcai/Sohrabi19,DBLP:journals/jair/Helmert06}.
Given a description of the possible initial states of the world, a description of the desired goals, and a description of a set of possible actions, the classical planning problem involves synthesizing a plan that, when applied to any initial state, generates a state which contains the desired goals (goal state)~\cite{DBLP:books/daglib/0014222}. The planning domain definition language (PDDL) serves as a standardized encoding of classical planning problems~\cite{Ghallab98,DBLP:series/synthesis/2019Haslum}. State of the art classical planning tools such as Pyperplan~\cite{alkhazraji-et-al-zenodo2020} and Fast Downward Planner~\cite{DBLP:journals/jair/Helmert06} remain far more efficient than LLM based end-to-end plan generation in terms of wall-clock time and accuracy of plan generation.

We address the problem of accurately responding to planning related queries by combining the strengths of LLMs, logical reasoners and classical planners. Fig.~\ref{fig:e2e-Diagram} illustrates a high level overview of this process. Some recent works have also investigated the use of LLMs and external planners. For example, \cite{xie2023translating} shows that LLMs are more adept at translation than planning, in particular,
by leveraging commonsense knowledge and reasoning to furnish missing details from under-specified goals (as is often the case in natural language). With this, LLMs can translate natural language task descriptions to input formats required by an external planner. The approach in~\cite{lyu2023faithful} presents preliminary results of integrating LLMs with PDDL on the SayCan dataset~\cite{ahn2022i}. The LLM+P approach~\cite{liu2023llmp} translates a natural language (NL) planning task description to task PDDL. However, while it yields better performance than relying solely on an LLM for end-to-end planning, it does not achieve acceptable accuracy for even slightly complex planning domains. One of the primary reasons for failure is that the LLM often makes errors generating information that must abide by the constraints specified in the domain knowledge or the task description. 

\begin{myfigure}[label={fig:example-task-description}]{Example Barman task description}
{\small \textit{You have 1 shaker with 3 levels, 5 shot glasses, 3 dispensers for 3 ingredients. The shaker and shot glasses are clean, empty, and on the table. Your left and right hands are empty. The first ingredient of cocktail1 is ingredient2. The second ingredient of cocktail1 is ingredient1. The first ingredient of cocktail2 is ingredient1. The second ingredient of cocktail2 is ingredient2. The first ingredient of cocktail3 is ingredient1. The second ingredient of cocktail3 is ingredient3. The first ingredient of cocktail4 is ingredient3. The second ingredient of cocktail4 is ingredient2. Your goal is to make 4 cocktails. shot1 contains cocktail1. shot2 contains cocktail4. shot3 contains cocktail3. shot4 contains cocktail2.}
}
\end{myfigure}
\vspace{0.5cm}
\begin{wrapfigure}{r}{0.5\textwidth}
\begin{tcolorbox}[left=0pt,right=0pt,top=0pt,bottom=0pt,size=fbox]
{\small
\begin{verbatim}
(define (problem prob)
 (:domain barman)
 (:objects 
  shaker1 - shaker
  shot1 shot2 shot3 shot4 
  shot5 - shot ...)
 (:init 
  (clean shaker1) (clean shot1)
  (clean shot2) (clean shot3) 
  (clean shot4) (clean shot5)
  ... (cocktail_part1 cocktail1
  ingredient2) ...
  (cocktail_part2 cocktail4
  ingredient2))
 (:goal (and
  (contains shot1 cocktail1)
  (contains shot2 cocktail4)
  (contains shot3 cocktail3)
  (contains shot4 cocktail2))))
\end{verbatim}
}
\end{tcolorbox}
\caption{Task PDDL of example task description in Fig.~\ref{fig:example-task-description}\vspace{-.5cm}}
\label{fig:example-task-pddl}
\end{wrapfigure}

Consider the planning task description in Fig.~\ref{fig:example-task-description} from the planning domain - \textit{Barman}~\footnote{Throughout this paper, we use the seven planning domains, namely, Barman, Blocksworld, Floortile, Grippers, Storage, Termes and Tyreworld. The natural language task descriptions of each domain along with the domain model PDDL can be found at~\url{https://github.com/Cranial-XIX/llm-pddl}. Task PDDLs for these and several other planning domains can be generated using the collection presented in~\cite{seipp_2022_6382174}.}. It contains cardinality information of shots (\textit{5 shot glasses}), but mentions only 4 shot instances explicitly. However, the correct task PDDL should contain 5 shot instances (see Fig.~\ref{fig:example-task-pddl}). An LLM does not always abide by this relationship. Similarly, the statement  \textit{all shots are clean} should lead to generation of PDDL atoms \texttt{(clean shot$i$)} with $1\leq i \leq 5$. But, an LLM sometimes generates only 4 such atoms (one for each mentioned shot instance). As another example, in the floortile domain, the user query only mentions the robots' position but the task PDDL should also contain information about which positions are clear. In this case, an LLM often fails to establish the logical relationship between occupied and clear positions.

Such logical dependency rules are deterministic and can be computed in an error-free manner with appropriate logic reasoners to augment the information extracted by LLMs. The approaches presented in~\cite{DBLP:journals/corr/abs-2101-11707,DBLP:journals/corr/abs-2302-03780,DBLP:conf/acl/YangI023} translate a user query to a structured representation and then uses domain rules to infer more information. However, they have shown to support only input sentences that directly translate to facts. As a result, they are not applicable for complex planning tasks which require support for translation to rules and constraints as well.

\begin{wrapfigure}{r}{0.5\textwidth}
  %\centering
    \vspace{-.6cm}
    \includegraphics[width=\linewidth]{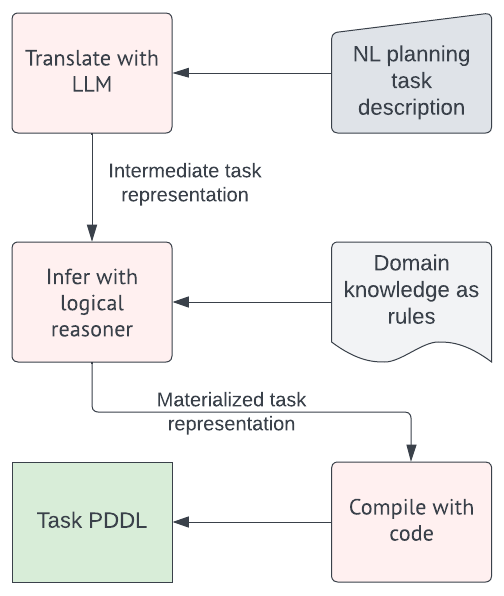}
    \vspace{-.5cm}
    \caption{TIC pipeline for generating task PDDLs from NL task descriptions\vspace{-.5cm}}
    \label{fig:TIC pipeline}
\end{wrapfigure}

The main idea underlying our approach is to combine the best of both these technologies (LLMs and logical reasoning) to achieve more accurate results for task PDDL generation, even for complex planning problems, than is feasible using either of them alone. More precisely, we break down the problem of task PDDL generation into the following three steps (see also Fig.~\ref{fig:TIC pipeline}).
\begin{enumerate}
\item \textbf{Translate} with an LLM, the natural language planning task description to a logical language with sufficient expressiveness to model facts, rules and constraints. We call such a logical representation an intermediate representation of the natural language planning task description.
\item \textbf{Infer}, with a reasoner for the chosen logical language, the rest of the required information using the intermediate representation and domain knowledge rules. We call the union of facts in the intermediate representation and the inferred facts as the materialized representation.
\item \textbf{Compile} the target task PDDL using the materialized representation and domain PDDL. 
\end{enumerate}
%Consequently, we refer to our approach as Translate-Infer-Compile (TIC) approach.

In this paper, we use Answer Set Programming (ASP)~\cite{DBLP:conf/iclp/GelfondL88,DBLP:conf/aaai/Lifschitz08} as the logical language. To the best of our knowledge, ours is the first approach that proposes an ASP based schema to represent complex queries - facts, constraints, complex data structures and rules. In Section~\ref{sec:asp-schema} we present the ASP schema for modeling planning tasks. Sections~\ref{sec:translate},~\ref{sec:infer}, and~\ref{sec:compile} present details of the translate, infer and compile steps respectively. In particular, Section~\ref{sec:translate} presents two techniques for translating a NL task description to an intermediate representation: (a) in-context based technique, and (b) generic domain-independent prompt based technique. Our experimental results presented in Section~\ref{sec:experiments} show that \begin{enumerate*}
\item TIC supports complex planning problems.
\item TIC achieves high accuracy (100\% in some cases) on all seven planning domains of the evaluation dataset presented in~\cite{liu2023llmp}.
\item TIC is also more general as it achieves high accuracy (near 100\% in some cases) on the same dataset but with variation in language of the task descriptions.
\end{enumerate*}

\section{ASP Schema for Task PDDLs}
\label{sec:asp-schema}
We first define the ASP schema for modeling the information in task descriptions (as intermediate representations) as well as that required in target task PDDLs (as materialized representations). In the following, \textbf{\textit{p/N}} stands for a predicate \textbf{\textit{p}} with arity (i.e., number of arguments) \textbf{\textit{N}}.

\paragraph{\textbf{object/2}} predicate relates objects with their respective types. For example, a fact \texttt{object(cocktail1, cocktail)} means that \texttt{cocktail1} is an object of type \texttt{cocktail}. 

\paragraph{\textbf{cardinality/2}} predicate represents the cardinality of the set of objects of a type. For example, \texttt{cardinality(shot, 5)} means that there are five shot glasses. Such cardinality information is often present in task descriptions whereas the task PDDL does not include this information but only the set of objects that satisfy the cardinality constraint. The following cardinality rules generate new objects of the given type when necessary to satisfy the cardinality constraints, where \texttt{make\_id} is a predicate external to an ASP reasoner.
\vspace{0.125cm}\\
\texttt{
\{object(@make\_id(X, T), T)\} :- cardinality(T, N), X = 1..N. \\
:- \#count {X: object(X, T)} = M, cardinality(T, N), M != N.
}
\vspace{0.125cm}\\
For example, given \texttt{cardinality(shot, 5)}, \texttt{object(shot1, shot)}, \texttt{object( \\
shot2, shot)}, \texttt{object(shot3, shot)}, and \texttt{object(shot4, shot)}, the cardinality rules above would generate the new fact \texttt{object(shot5, shot)}. %\todo{explain why N+1 ... 2N}.

\paragraph{\textbf{init/1} and \textbf{goal/1}} predicates differentiate initial state facts from goal state facts. All domain predicates defined in the domain.pddl, the map/3 and grid predicates (described below) may occur as arguments of \texttt{init} and \texttt{goal}. A task description describes the init and goal states as a set of facts that are true in the respective states. For example, an initial state sentence \textit{shot3 is on the table} is represented as \texttt{init(ontable(shot3))}, and a goal state sentence \textit{shot3 should be on the table} is represented as \texttt{goal(ontable(shot3))}.

\paragraph{\textbf{map/3}} predicate is used as a shorthand for a bijective relationship between two objects types (of the same cardinality). For example, \texttt{map(dispenser, \\
dispenses, ingredient)} stands for \texttt{dispenses(dispenser1, ingredient1)}, 
\ldots, \\       
   \texttt{dispenses(dispenserN, ingredientN)}, where \texttt{N} is the cardinality of type \\ \texttt{dispenser} as well as of type \texttt{ingredient}. The rules below expand a map shorthand to corresponding facts of the initial state (analogous rules for the goal state) where \texttt{make\_fact} is a predicate external to an ASP reasoner.
\vspace{0.125cm}\\   
\texttt{
1 {init(@make\_fact(X, P, Y)) : object(X, T1)} 1 :-\\       
  init(map(T1, P, T2)), object(Y, T2). \\
1 {init(@make\_fact(X, P, Y)) : object(Y, T2)} 1 :-\\   
 init(map(T1, P, T2)), object(X, T1).
}

\paragraph{\textbf{Two-dimensional Grid.}} Some planning domains require capturing the positions of certain objects on a two-dimensional grid. For example, in the Floortile domain, a tile \texttt{tile\_x\_y} means that the tile is on row \texttt{x} and column \texttt{y}. We represent such information with a predicate \textbf{grid/3} where grid is a placeholder for the appropriate grid name. For example, \texttt{floortile\_grid(x, y, tile\_x\_y)} represents the position of Floortile tile object \texttt{tile\_x\_y}.

Note that some planning domains may require extensions to this schema for supporting other data structures that we haven’t implemented yet. We believe that ASP is expressive enough to support such extensions.

\section{Translation to Intermediate Representation}
\label{sec:translate}
In this section, we introduce intermediate representation, the core idea underlying the TIC approach, and present two approaches for generating the intermediate representation of a given task description. The first approach uses an in-context example for each domain whereas the second approach uses a generic domain independent prompt that works for all domains.

\subsection{Intermediate Representation}

\begin{wrapfigure}{r}{0.5\textwidth}
 \vspace{-0.9cm}
\begin{tcolorbox}[left=0pt,right=0pt,top=0pt,bottom=0pt,size=fbox]
{\small
\begin{verbatim}
cardinality(shaker, 1).
cardinality(level, 3).
cardinality(shot, 5).
cardinality(dispenser, 3).
cardinality(ingredient, 3).
init(clean(X)) :- object(X, 
 shaker).
init(clean(X)) :- object(X, shot).
...
init(handempty(left)).
init(handempty(right)).
init(map(dispenser, dispenses, 
    ingredient)).
init(cocktail_part1(cocktail1,
    ingredient2)).
...
init(cocktail_part2(cocktail4,
    ingredient2)).
goal(contains(shot1, cocktail1)).
...
goal(contains(shot4, cocktail2)).
\end{verbatim}
}
\end{tcolorbox}
\caption{Intermediate representation of the example task description in Fig.~\ref{fig:example-task-description}}
\vspace{-0.4cm}
\label{fig:example-intermediate-representation}
\end{wrapfigure}

An intermediate representation of a query is a logically interpretable representation of the task description. Intermediate representations capture information that is directly present in the query, while additional domain knowledge generates missing (and deterministically inferable) information. Intermediate representations serve multiple purposes to address the accuracy bottleneck of LLMs when generating the task PDDL directly.

%Unlike the in-context examples of direct PDDL generation approaches, an intermediate representation does not contain information that is not present in the task description but required in the task PDDL. In some cases, the missing information can be inferred from the information in intermediate representation. For example, if a barman task description states that there are 5 shots as well as mentions \texttt{shot1}, \texttt{shot2}, \texttt{shot3}, and \texttt{shot4}, then the intermediate representation should not contain inferable information \texttt{shot5}. In other cases, the missing information can be added with the help of rules derived from the domain knowledge. For example, a barman task description may not contain information corresponding to the PDDL facts \texttt{(next l0 l1)}, \texttt{(next l1 l2)}, \texttt{(shaker\_empty\_level shaker1 l0)}, and \texttt{(shaker\_level shaker1 l0)} because the knowledge that the shaker levels have an order is domain knowledge that is common for all barman queries.

An intermediate representation facilitates bridging syntactic and semantic heterogeneities between the task description and target task PDDL. For example, both the sentences \textit{You have 3 shots; all shots are clean.}, and \textit{Shot 1, shot 2, and shot 3 are clean.} should be translated to the PDDL facts \texttt{(clean shot1)}, \texttt{(clean shot2)} and \texttt{(clean shot3)}, in target task PDDL. In the first case, the intermediate representation contains the rule \texttt{init(clean(X)) :- object(X, shot)}. In the second case, the intermediate representation contains the facts \texttt{init(clean(shot1))}, \texttt{init(clean(shot2))}, and \texttt{init(clean(shot3))}. In both cases, a logical reasoner infers a model that contains the facts \texttt{init(clean(shot1))}, \texttt{init(clean(shot2))}, and \texttt{init(clean(shot3))}, thus bridging the syntactic gap. In another case, a given domain.pddl may not have semantically meaningful object types and predicate names. For example, the Blocksworld domain does not define the object type \texttt{block} but requires the blocks to be instances of the generic object type \texttt{object}. In such cases, an intermediate representation uses the object types and predicates that are closer to natural language than to those of the target PDDL. A logical reasoner maps objects of type \texttt{block} to objects of type \texttt{object} using a simple rule \texttt{object(X, object) :- object(X, block)}.

Some sentences in the task description are more naturally and compactly represented intensionally as a rule rather than extensionally by enumerating all implied facts. For example, the phrase \textit{3 dispensers for 3 ingredients} is more directly represented as the shortcut \texttt{map(dispenser, dispenses, ingredient)} rather than by explicitly enumerating all the facts. In another example, the initial state sentence \textit{The shakers are empty} is represented with the rule \texttt{init(empty(X)) :- object(X, shaker)} rather than with an enumeration of all implied facts.

As a result, when an LLM generates an intermediate representation of a task description, the LLM output is more often correct. The LLM output is also shorter and doesn’t grow if the query contains a lot of derived information. In addition, we require shorter in-context examples thus leading to shorter prompts. %In Section~\ref{sec:nl-to-int-rep} we present two approaches for translating natural language planning task descriptions to intermediate representations.

%\subsection{Translating NL Task Description to Intermediate Representation using LLM}
%\label{sec:nl-to-int-rep}
%In this section, we present two approaches for generating the intermediate representation of a given task description. The first approach uses an in-context example for each domain whereas the second approach uses a generic prompt that works for all domains.

\subsection{Using In-Context Domain Examples}
%LLMs are known for their proficiency in learning in context without the need for fine-tuning their parameters. 
In-context learning in LLMs is their ability to handle unseen tasks using only a few demonstrations represented as input-label pairs ~\cite{brown2020language}. In the in-context example based approach, we use an LLM with a domain specific in-context example to generate the intermediate representation of a given task description. The LLM prompt consists of an example task description, its intermediate representation, and the new task task description as shown in Fig.~\ref{fig:in-context-example-prompt}. We refer to this approach as TIC-IC. The complete prompt for the barman domain is presented in Appendix~\ref{appendix:in-context-prompts}.  Fig.~\ref{fig:example-intermediate-representation} shows the intermediate representation of the example task description in Fig.~\ref{fig:example-task-description} generated by an LLM using TIC-IC.

\begin{myfigure}[label={fig:in-context-example-prompt}]{LLM prompt template for generating intermediate representation using domain specific in-context example}
{I want you to create ASP Logic Program representation of a paragraph.
\\
\\
An example paragraph is: \textit{<example task description>}
\\
\\
The ASP Logic Program representation of the example paragraph is:
\textit{<intermediate rep of example task description>}
\\
\\
Now I have a new paragraph: \textit{<new task description>}
\\
\\
Provide me with the ASP Logic Program representation of the new paragraph directly without further explanations.}

\end{myfigure}

\subsection{Using Generic Prompts}
The approach based on in-context examples achieves high accuracy but requires an in-context example for each domain. This can be a bottleneck in use cases that need to support many, possibly previously unknown, types of user queries. In this subsection, we present a domain independent generalized prompt that addresses this limitation. The approach consists of three steps:
\begin{enumerate*}
    \item Cardinality Extraction
    \item Named Objects Extraction
    \item Rules Extraction
\end{enumerate*}. The examples used in the prompts for these steps are from an imaginary domain that is not part of our test dataset. The examples domain is sufficiently expressive in the sense that it includes all constructs and data structures needed for modeling planning problems in our test dataset domains. Thus, the constructed prompts are domain independent.

\paragraph{\textbf{Cardinality Extraction:}}
This step extracts the explicitly mentioned cardinalities of object types in the given task description. The LLM prompt for this step consists of instructions to \begin{enumerate*}
    \item split the task description into initial state and goal state, and
    \item extract for each object type $T$, the count of $T$ that is explicitly mentioned in the initial state description
\end{enumerate*} (see Appendix~\ref{appendix:generic-prompts} for complete prompt). The set of provided object type definitions includes a name and a short description for each domain object type. For the example task description in Fig.~\ref{fig:example-task-description}, the LLM generates the following output:
{\texttt{
\{shaker: 1, level: 3, shot: 5, dispenser: 3, hand: 2, ingredient: 3\}
}}

\paragraph{\textbf{Named Objects Extraction:}}
This step extracts the explicitly mentioned named objects for each object type in the given task description. The LLM prompt for this step consists of instructions for detecting and formatting object instances (see Appendix~\ref{appendix:generic-prompts} for complete prompt). The set of provided object types is the same as that in the previous step. For the example task description in Fig.~\ref{fig:example-task-description}, the LLM generates the following output:
{
\texttt{
\{shaker: [], level: [], shot: [shot1, shot2, shot3, shot4], dispenser: [], 
hand: [left, right], \\
cocktail: [cocktail1, cocktail2, cocktail3, cocktail4], \\
ingredient: [ingredient1, ingredient2, ingredient3]\}
}
}

\paragraph{\textbf{Rules Extraction:}}
In this step, the LLM generates a Logic Program representation of the given task description. The LLM prompt consists of steps to detect the initial and goal states, translate each state description per state translation rules, and output the result as syntactically correct ASP Logic Program. The LLM is instructed to use both the provided set of extracted objects as well as the set of domain predicates. The latter consists of the predicate's name, arity, argument types, and a brief description. (see Appendix~\ref{appendix:generic-prompts} for example object types and predicates). The following describe the state translation rules where \texttt{<state>} is a placeholder for \texttt{init} or \texttt{goal}.
\begin{enumerate}
    \item A property of all instances of an object type. For example, \textit{All apples are clean} should be translated as \texttt{<state>(clean(X)) :- object(X, apple)}.
    \item A property of a named object. For example, \textit{apple1 is red} should be translated as \texttt{<state>(red(apple1))}.
    \item A relationship between two or more named objects. For example, \textit{apple1 is on plate2} should be translated as \texttt{<state>(on(apple1, plate2))}.
    \item A relationship between two or more unnamed objects. For example, \textit{cups' handles are free} should be translated as \texttt{<state>(free(X, Y)) :-\\
    cup\_handle(X, Y)}.
    \item A relationship between all unnamed objects of an object type with a named object. For example, \textit{apples are on table1} should be translated as \\
    \texttt{<state>(on(X, table1)) :- object(X, apple)}.
    \item A bijective relationship between instances of two object types of the same cardinality. For example, \textit{3 balls for 3 kids} should be translated as \texttt{<state> \\
    (map(ball, play, kid))}.
    \item A 2 dimensional grid of cells with coordinates. For example, \textit{cell\_1\_1, \\
    cell\_1\_2, cell\_2\_1, cell\_2\_2} should be translated as \texttt{<state>(cell\_grid\\
    (1, 1, cell\_1\_1))}, \ldots, \texttt{<state>} \texttt{(cell\_grid(2, 2, cell\_2\_2))}.
\end{enumerate}
For the example task description in Fig.~\ref{fig:example-task-description}, the LLM generates the same intermediate representation as in Fig.~\ref{fig:example-intermediate-representation}.

The steps for extracting cardinalities, named objects and rules can be performed with three separate LLM calls (referred to as TIC-G3), or together in one LLM call (referred to as TIC-G1). While extracted named objects are programmatically passed on to the next prompt for rule extraction in TIC-G3, in TIC-G1, the LLM is instructed to extract named objects and subsequently use them for rule extraction. Refer to Appendix~\ref{appendix:generic-prompts} for complete LLM prompts of both implementation options.

\section{Inference of Materialized Representation}
\label{sec:infer}
An intermediate representation contains only the information present in a task description. We compute the remaining information required in the task PDDL deterministically using an ASP solver. A materialized representation is the union of the facts in the intermediate representation and the inferred facts. In this section, we present different kinds of ASP rules that may be relevant for inferring information in a planning domain. Note that not all kind of rules may be required for every planning domain.

1. Rules to infer the type of objects that need to be declared in the task PDDL. For object types that are arguments of a predicate, we can automatically generate the rules for deriving the object types from the predicate definitions in domain.pddl. For example, from the predicate definition \texttt{(cocktail\_part1 ?c - cocktail ?i - ingredient)} we can generate the rules:
\vspace{0.125cm}\\
\texttt{
object(X, cocktail):- cocktail\_part1(X, \_).\\
object(X, ingredient):- cocktail\_part1(\_, X).
}
\vspace{0.125cm}\\
In cases where there is no predicate to infer a certain object type, one has to add the corresponding rules manually. %This case doesn't occur in the Barman domain. However, 
For example, a Blocksworld task PDDL requires the blocks to be of type \texttt{object}, while they are of type \texttt{block} in the intermediate representation. Therefore, we require a rule \texttt{object(X, object) :- object(X, block).} %to assign the type \texttt{object} to each block.

The following two kinds of rules can be authored semi-automatically based on domain knowledge that may already be present in the domain.pddl, or some documentation, or with the help of domain experts. 

2. Rules to infer information derivable from other information present in the query. For example, for the Floortile domain, the rule \texttt{init(clear(T)) :- object(T, tile), not init(robot\_at(\_, T))} infers that a tile is clear if there is no robot at that tile.

3. Rules to infer information that is not derivable from information present in the query. For example, in the Barman domain, the task PDDL requires the order of shaker levels known through domain knowledge. The following rules
\vspace{0.125cm}\\
\texttt{
init(@make\_seq(N-1, l, next, 1)) :- cardinality(level, N).\\
first\_level(X) :- object(X, level), not init(next(\_, X)).    
}
\vspace{0.125cm}\\
\noindent generate the facts \texttt{init(next(level0, level1))}, \texttt{init(next(level1, \\level2))}, and \texttt{first\_level(level0)} for \texttt{N=3}, where \texttt{make\_seq} is a predicate for generating a sequence and is external to ASP. See Appendix~\ref{sec:domain-rules} for complete listing of rules for all seven planning domains considered in this work.

\section{Compilation to Target Representation}
\label{sec:compile}
In the last step, we generate the task PDDL. The materialized representation contains all the information necessary in a task PDDL. So, this step is mostly a straightforward syntactic transformation of facts from the ASP Logic Program syntax to the PDDL syntax. In principle, this step can also be performed using an LLM. But, since the translation process is deterministic and simple, it is less likely to encounter errors when performed through code. The code based transformation consists of three main steps: 
\begin{enumerate}
    \item For all bindings \texttt{(obj1, t), \ldots, (objn, t)} in the answer to the ASP query \texttt{object(X, T)}, if object type \texttt{t} is defined in the domain.pddl, generate a PDDL string "\texttt{obj1 \ldots\  objn - t}". The concatenation of such strings makes the objects section of the task PDDL.
    \item For all bindings \texttt{p(obj1, \ldots, objn)} of \texttt{X} in the answer to the ASP query \texttt{init(X)}, if the predicate \texttt{p} is defined in the domain.pddl, then generate a PDDL string "\texttt{(p obj1 \ldots\ objn)}". The concatenation of such strings makes up the init section of the task PDDL.
    \item Similar to the previous step, generate the PDDL string for all bindings of \texttt{X} for the query \texttt{goal(X)}. The string representing the conjunction of all such strings makes up the goal section of the task PDDL.
\end{enumerate}

\section{Experiments}
\label{sec:experiments}
Through experiments we address the following questions. 1. Does introducing an intermediate representation and a logical reasoner improve the accuracy of PDDL generation? 2. Does the TIC approach generalize to language variations in the natural language task description? 3. How does the LLM perform with generic translation instructions for intermediate representation?

\subsection{Experiment Setup}
We evaluate our TIC approach on seven planning domains that have been taken directly from LLM+P~\cite{liu2023llmp} and are standard planning benchmarks~\footnote{This test dataset is based on a collection that has been used for IPC}. Each of the domains include a domain description PDDL and 20 automatically generated planning tasks from~\cite{seipp_2022_6382174}, with each task consisting of a natural language description and a corresponding ground truth PDDL. For evaluating the in-context example based approach, we also include an example task description and a corresponding intermediate representation that serves as the in-context example for each domain. In addition, we also created a new dataset derived from the LLM+P dataset by introducing language variation in each of the natural language task descriptions for the seven domains. We used GPT-3.5 to paraphrase the original task descriptions and call this the \textit{Language Variation} dataset. We evaluate the TIC approach on the Language Variation dataset without modifying the in-context learning examples or the generic prompt. We use Clingo~\cite{DBLP:journals/tplp/GebserKKS19} python package\footnote{https://pypi.org/project/clingo/} as the ASP solver. For each of the tasks, we generate the task PDDL and compare it with the ground truth PDDL using an automated process. Note that the task PDDL comparison algorithm does not count the number of mismatches but determines whether two PDDLs are equivalent or not (see Appendix~\ref{appendix:comparison} for details). %We solely present the PDDL generation accuracy since plan generation using a classical planner is deterministic and precise provided that the PDDL is accurate for the given dataset. 
Our experiments evaluate below listed approaches on GPT-4, GPT-3.5 Turbo and PaLM2 (chat-bison) models for both the LLM+P as well as the Language Variation datasets. 

\begin{itemize}
    \item LLM+P: The LLM+P approach of directly (i.e. without use of intermediate representations or logical reasoning) generating task PDDL using an LLM, as a baseline. This baseline has been shown to outperform LLM as a planner as well as LLM + external planner without in-context examples.
    \item TIC-IC: Our TIC In-Context example based approach; requires one LLM call.
    \item TIC-G3: Our TIC Generic Prompt based approach with three LLM calls.
    \item TIC-G1: Our TIC Generic Prompt based approach with one LLM call.
\end{itemize}

\vspace{-.7cm}
\begin{table*}[ht]
\centering
    \begin{tabular}{|c|c c c c|c c c c|}
    \hline
    {\textbf{Planning}} & \multicolumn{4}{c|}{\textbf{LLM+P Dataset}} & \multicolumn{4}{c|}{\textbf{Language Variation Dataset}} \\ 
    {\textbf{Domain}} &\, LLM+P & TIC-IC & TIC-G1 & TIC-G3 \, & \, LLM+P & TIC-IC & TIC-G1 & TIC-G3 \,\\
    \hline
    {} & \multicolumn{8}{|c|}{\textbf{GPT-4}}\\
    \hline
    Barman & \textbf{100} &  \textbf{100} & \textbf{100} & 98.33 & \textbf{100} & \textbf{100} & \textbf{100} & 96.67\\
    \hline
    Blocksworld & 88.33  & \textbf{100} & \textbf{100} & \textbf{100} & \textbf{100} & \textbf{100} & \textbf{100} & \textbf{100}\\
    \hline
    Floortile & 0  & \textbf{100} & \textbf{100} & \textbf{100} & 0 & \textbf{100} & \textbf{100} & \textbf{100}\\
    \hline    
    Grippers & 95  & \textbf{100} & \textbf{100} & 98.33 & 0 & \textbf{100} & 96.67 & 96.67\\
    \hline    
    Storage & 40  & \textbf{100} & \textbf{100} & 98.33 & 0 & \textbf{100} & 93.33 & 68.33\\
    \hline
    Termes & 53.33  & \textbf{100} & 98.33 & 98.33 & 0 & \textbf{100} & \textbf{100} & 96.67\\
    \hline    
    Tyreworld & 71.67  & \textbf{100} & 95 & 95 & 0 & \textbf{100} & 93.33 & 88.33\\
    \hline

    {} & \multicolumn{8}{|c|}{\textbf{GPT-3.5}}\\
    \hline
    Barman & 98.33 & \textbf{100} & 0 & 0 & 53.33 & \textbf{100} & 0 & 0\\
    \hline
    Blocksworld & 71.67 & \textbf{100} & 91.67 & 93.33 & 51.67 & \textbf{95} & 55 & 63.33\\
    \hline
    Floortile & 26.67 & \textbf{100} & \textbf{100} & 68.33 & 0 & \textbf{100} & 83.33 & 48.33\\
    \hline
    Grippers & 95 & \textbf{100} & 40 & 36.67 & 0 & \textbf{90} & 15 & 35\\
    \hline
    Storage & 25  & \textbf{100} & 0 & 0 & 0 & \textbf{80} & 0 & 0\\
    \hline
    Termes & 71.67 & \textbf{100} & 15 & 63.33 & 5 & \textbf{100} & 0 & 21.67\\
    \hline
    Tyreworld & 55 & \textbf{100} & 8.33 & 0 & 0 & \textbf{90} & 0 & 0\\
    \hline

    {} & \multicolumn{8}{|c|}{\textbf{PaLM2}}\\
    \hline
    Barman & 75 & \textbf{100} & \textbf{100} & 80 & 0 & \textbf{100} & 66.67 & 66.67\\
    \hline
    Blocksworld & 10 & \textbf{100} & \textbf{100} & \textbf{100} & 0 & \textbf{100} & 95 & 95\\
    \hline
    Floortile & 0 & \textbf{90} & \textbf{90} & 60 & 0 & \textbf{90} & \textbf{90} & 45\\
    \hline
    Grippers & 95 & \textbf{100} & 95 & 95 & 0 & \textbf{100} & 60 & 60\\
    \hline
    Storage & 0 & \textbf{100} & 60 & 6.67 & 0 & \textbf{100} & 50 & 0\\
    \hline
    Termes & 0 & \textbf{100} & 93.33 & 0 & 0 & \textbf{100} & 65 & 0\\
    \hline
    Tyreworld & 35 & \textbf{100} & 0 & 0 & 0 & \textbf{95} & 0 & 0\\
    \hline    
    \end{tabular}
    \vspace{0.25cm}
    \caption{\label{table:Results3} \% Accuracy of PDDL generation on Language Variation dataset averaged over 3 runs (variance $\leq 1$)\vspace{-1.25cm}}    
\end{table*}

\subsection{Results and Analysis}
Table~\ref{table:Results3} presents our experiment results. TIC-IC outperforms LLM+P approach on both the LLM+P dataset and the Language Variation dataset in all seven domains. In particular, TIC-IC achieves 100\% accuracy on the LLM+P dataset even for complex domains such as Floortile, Storage and Termes with GPT-3.5 Turbo and GPT-4. This highlights TIC's ability to generate highly accurate task PDDLs using logically interpretable intermediate representations of the natural language task descriptions. The reason for slightly lower accuracy of TIC-IC in Floortile with PaLM2 is that the chat-bison model truncates the output of 2 out of 20 problems. We also observe that TIC-IC results in a high accuracy on the Language Variation dataset with GPT-4 and PaLM2. 

In general, the lower accuracy of the TIC approaches on the Language Variation dataset compared to the LLM+P dataset is due to more complex and longer sentences in the task descriptions of the former dataset. See Appendix~\ref{appendix: language variation example} for examples. 

As we move on to the generic prompt based approach, TIC-G3 generalizes well overall on GPT-4. It achieves $ \geq95\%$ accuracy on all seven domains (with 100\% on five domains) of the LLM+P dataset and $ \geq93.33\%$ on all seven domains (with 100\% on four domains) of the Language Variation dataset. In general, the lower accuracy of TIC-G1 as compared to TIC-G3 may be due to the more complex instructions to LLM for extracting cardinalities, objects and rules in a single step. We also observe a drop in accuracy of TIC-G3 and TIC-G1 for some domains on PaLM2 and GPT-3.5 models when compared to GPT-4. This may be due to the failure of PaLM2 and GPT-3.5 in abiding by the generic instructions in some cases. Some of the common errors by these LLMs are in extracting bijective maps and rules as well as in generating syntactically correct ASP rules.

In our setup, one LLM call takes 3-5 seconds. Overall, more than 99\% of total time for processing a query is consumed by the LLM call(s) in the translation step.

\section{Conclusion}
We investigated the problem of answering natural language planning task requests by combining LLMs with an external planner. We presented the Translate-Infer-Compile (TIC) approach to automatically generate highly accurate task PDDLs from the task requests. %TIC addresses the limitations of LLMs in generating accurate plans in the presence of a large number of actions by enabling use of external planners. 
The core idea of TIC lies in the introduction of a logically interpretable intermediate representation of the task description. TIC combines the strengths of LLMs and symbolic reasoning to address LLMs' limitation in directly generating task PDDLs from task descriptions. %Further, TIC enables the use of an external planner to address LLM's limitation of accurate plan generation. 
TIC not only outperforms similar approaches with respect to the accuracy of task PDDL generation but is also more general.

Although, in this work, we have focused on generating planning task PDDLs, the TIC approach is generic enough to be applicable for use cases that need to support a natural language interface for structured tasks such as mathematical problems (see also Tool-integrated Reasoning Agents (ToRA)~\cite{gou2023tora}), API calls (see also ToolFormer~\cite{schick2023toolformer}, LangChain~\cite{langchain}, and Gorilla~\cite{patil2023gorilla}), and database queries (see also Spider dataset~\cite{DBLP:journals/corr/abs-1809-08887} and the benchmark accuracy on SQL databases~\cite{sequeda2023benchmark}). 

While the ability to do logical reasoning helps us achieve high accuracy, it comes with a price of modeling the domain. However, since the domain modeling is an offline task, we believe that it can be accelerated with semi-automatic techniques, possibly using LLMs among others, e.g.,~\cite{DBLP:conf/kr/IshayY023}. In this context, we believe that Description Logics~\cite{DBLP:conf/dlog/2003handbook} as a formalism and the standard Web Ontology Language (OWL)~\cite{owl2-overview} may lend themselves as alternatives to ASP. 


\bibliographystyle{splncs04}
\bibliography{paper}

%
% ---- Appendix ------
%\begin{subappendices}
%\renewcommand{\thesection}{\Alph{section}}%
\clearpage
\appendix

\section{LLM Prompt for Generating Intermediate Representation using In-Context Example}
\label{appendix:in-context-prompts}

\begin{tcolorbox}[enhanced, breakable, title=LLM Prompt for the Barman domain for generating intermediate representation using domain specific in-context example,size=fbox]
I want you to create ASP Logic Program representation of a paragraph.
\\
\\
An example paragraph is: \textit{You have 1 shaker with 3 levels, 4 shot glasses, 3 dispensers for 3 ingredients. The shaker and shot glasses are clean, empty, and on the table. Your left and right hands are empty. The first ingredient of cocktail1 is ingredient3. The second ingredient of cocktail1 is ingredient1. The first ingredient of cocktail2 is ingredient1. The second ingredient of cocktail2 is ingredient2. Your goal is to make 3 cocktails. shot1 contains cocktail1. shot2 contains cocktail2. shot3 contains cocktail1.}
\\
\\
The ASP Logic Program representation of the example paragraph is:
\begin{verbatim}
cardinality(shaker, 1).
cardinality(level, 3).
cardinality(shot, 4).
cardinality(dispenser, 3).
cardinality(ingredient, 3).
cardinality(cocktail, 2).
init(clean(X)) :- object(X, shaker). 
init(clean(X)) :-  object(X, shot).
init(empty(X)) :- object(X, shaker).
init(empty(X)) :- object(X, shot).
init(ontable(X)) :- object(X, shaker). 
init(ontable(X)) :- object(X, shot).
init(handempty(left)).
init(handempty(right)).
init(map(dispenser, dispenses, ingredient)).
init(cocktail_part1(cocktail1, ingredient3)).
init(cocktail_part2(cocktail1, ingredient1)).
init(cocktail_part1(cocktail2, ingredient1)).
init(cocktail_part2(cocktail2, ingredient2)).
goal(contains(shot1, cocktail1)).
goal(contains(shot2, cocktail2)).
goal(contains(shot3, cocktail1)).

\end{verbatim}
Now I have a new paragraph: \textit{
You have 1 shaker with 3 levels, 5 shot glasses, 3 dispensers for 3 ingredients. 
The shaker and shot glasses are clean, empty, and on the table. Your left and right hands are empty. 
The first ingredient of cocktail1 is ingredient2. The second ingredient of cocktail1 is ingredient1. 
The first ingredient of cocktail2 is ingredient1. The second ingredient of cocktail2 is ingredient2. 
The first ingredient of cocktail3 is ingredient1. The second ingredient of cocktail3 is ingredient3. 
The first ingredient of cocktail4 is ingredient3. The second ingredient of cocktail4 is ingredient2. 
Your goal is to make 4 cocktails. 
shot1 contains cocktail1. shot2 contains cocktail4. shot3 contains cocktail3. shot4 contains cocktail2. 
}
\\
\\
Provide me with the ASP Logic Program representation of the new paragraph directly without further explanations.
\end{tcolorbox}

\section{Generating Intermediate Representation using Generic LLM Prompts}
\label{appendix:generic-prompts}

\begin{tcolorbox}[enhanced, breakable, title=Text, size=fbox]
\textit{You have 2 tables, 5 plates, 4 apples, and 3 balls for 3 kids.
apple1 is on plate\_2. apple2 is on plate\_1. orange1 is on plate\_1. orange3 is on plate\_2. ball1 is in first drawer of first table. ball3 is in second drawer of first table. apple3 is in first drawer of second table. apple2 is in second drawer of second table. 
The apples are red. There are three oranges. The plates are clean. The kids are inside the hall.
cell\_0\_1 cell\_0\_2
cell\_1\_1 cell\_1\_2
cell\_2\_1 cell\_2\_2
cell\_3\_1 cell\_3\_2
There are 8 cells, each cell has a robot.
Your goal is to move the oranges to empty plates.
orange1 should be on plate\_3, orange2 should be on plate\_4,
orange3 should be on plate\_5. No orange should be unplated.
The plates are not clean. Each kid should paint a ball. The kids are inside the hall.
}
\end{tcolorbox}

\begin{tcolorbox}[enhanced, breakable, title=Object types,size=fbox]
\begin{verbatim}
[
 {'type': 'apple', 'description': 'The set of apples.'},
 {'type': 'orange', 'description': 'The set of oranges.'},
 {'type': 'plate', 'description': 'The set of plates.'},
 {'type': 'table', 'description': 'The set of tables.'},
 {'type': 'ball', 'description': 'The set of balls.'},
 {'type': 'kid', 'description': 'The set of kids.'}, 
 {'type': 'room', 'description': 'The set of rooms.'},
 {'type': 'cell', 'description': 'The set of cells.'},
 {'type': 'robot', 'description': 'The set of robots.'}
]
\end{verbatim}
\end{tcolorbox}

\begin{tcolorbox}[enhanced, breakable, title=Predicates,size=fbox]
\begin{verbatim}
[
{'predicate': 'on', 'arity': 2, 'argument types': ['apple', 
'plate'], 'description': 'on(a, b) means apple a is on 
plate b.'},

{'predicate': 'red', 'arity': 1, 'argument types': ['apple'],
'description': 'red(a) means apple a is red.'}, 

{'predicate': 'clean', 'arity': 1, 'argument types': 
['plate'], 'description': 'clean(a) means plate a is clean.'},

{'predicate': 'not_clean', 'arity': 1, 'argument types':
['plate'], 'description': 'not_clean(a) means plate a is 
not clean.'},

{'predicate': 'empty', 'arity': 1, 'argument types': 
['plate'], 'description': 'empty(a) means plate a is empty. A
plate is empty if there is no apple on it.'},

{'predicate': 'cell_grid', 'arity': 3, 'argument types': ['row', 
'column', 'cell'],  'description': 'cell_grid(a, b, c) means
cell a is on row b and column c.'},

{'predicate': 'unplated', 'arity': 0, 'argument types': [],
'description': 'unplated means at least one orange is not
on the plate.'},

{'predicate': 'play', 'arity': 2, 'argument types': ['ball',
'kid'], 'description':  'play is a bijective map between balls
and kids. play(a, b) means ball a is played with by kid b.'},

{'predicate': 'cell_robot_map', 'arity': 2, 'argument types': 
['cell', 'robot'], 'description': 'cell_robot_map is a bijective 
map between cells and robots.  cell_robot_map(a, b) means 
cell a has robot b.'},

{'predicate': 'in_1', 'arity': 2, 'argument types': ['apple or
ball', 'table'], 'description': 'in_1(a, b) means apple or ball
a is in first drawer of table b.'},

{'predicate': 'in_2', 'arity': 2, 'argument types': ['apple or
ball', 'table'],  'description': 'in_2(a, b) means apple or ball
a is in second drawer of table b.'},

{'predicate': 'inside', 'arity': 2, 'argument types': ['kid',
'room'], 'description': 'inside(a, hall) means kid a is
inside hall.'}
]
\end{verbatim}
\end{tcolorbox}

\begin{tcolorbox}[enhanced, breakable, title=LLM Prompt for Cardinality Extraction,size=fbox]
Given a list of object types and their descriptions as JSON. The provided text describes a planning task. The planning task describes an init state and a goal state. I want you to extract information from the provided task description by following below steps:\\
Step 1. Detect the init state description and the goal state description. The init state description is the text from start to begin of goal description. The goal description typically starts with 'Your goal is ...'. \\
Step 2. Extract for each object type T the explicitly mentioned count, width, and height of instances of T from the init state description.
\\
\\
Object types: [AS ABOVE]
\\
\\
Text : [AS ABOVE]
\\
\\
Answer: 
\begin{verbatim}
{'table': 2, 'apple': 4, 'orange': 3, 'plate': 5, 'ball': 3,
'kid': 3, 'cell': 8}

\end{verbatim}
Object types: \textlangle Task domain object types\textrangle
\\
\\
Text: \textlangle Task description\textrangle
\\
\\
Answer: 
\end{tcolorbox}

\begin{tcolorbox}[enhanced, breakable, title=LLM Prompt for named objects extraction,size=fbox]
Given a list of object types with their descriptions as JSON. 
Extract the instances of each object type from the provided text. Follow below
steps in order to do so.\\
Step 1. Instances appear as singular nouns in text. Ignore plural nouns.\\
Step 2. Consider only the instances that are explicitly mentioned in the provided text. Do not output instances that are not mentioned in the text.\\
Step 3. If an instance id contains spaces or hyphens, replace them by underscores.\\
Step 4. If an instance id begins with an upper case, then convert it to lower case.\\
\\
\\
Object types: [AS ABOVE]
\\
\\
Text: [AS ABOVE]
\\
\\
Answer:
\begin{verbatim}
{'apple': ['apple1', 'apple2'], 'orange': [], 'plate': 
['plate_1', 'plate_2',  'plate_3', 'plate_4', 'plate_5'], 
'table': ['table1', 'table2'], 'ball': ['ball1',  'ball3'],  
'room': ['hall'], 'kid': [], 'robot': [], 'cell': ['cell_0_1', 
'cell_0_2', 'cell_1_1', 'cell_1_2', 'cell_2_1', 'cell_2_2',
'cell_3_1', 'cell_3_2']}    
\end{verbatim}

Object types: \textlangle Task domain object types\textrangle
\\
\\
Text: \textlangle Task description\textrangle
\\
\\
Answer:
\end{tcolorbox}

\begin{tcolorbox}[enhanced, breakable, title=LLM Prompt for Rules Extraction,size=fbox]
Given
\\
(1) A JSON with object types as keys and list of objects as values.
\\
(2) A list of predicates. The predicates of arity one represent properties of objects. The predicates of arity two or more represent relationships of two or more objects.
\\
(3) A text describing a planning task. A planning task description consists of initial state and goal state descriptions. 
\\
\\
Translate the provided text following below steps:
\\
\\
Step 1. Extract the initial state and goal state descriptions.
\\
\\
Step 2. Translate each state description by following below rules as applicable.
'\textlangle state\textrangle' is a placeholder for 'init' or 'goal' for sentences of initial state or goal state respectively. While doing so, use only the provided predicates and respect their arity and type of arguments.
\\
\\
Step 3. Output the result as a valid ASP Logic Program without any comments or additional text.
\\
\\
State sentence translation rules:\\
Rule 1. Represent properties of sets of unnamed objects of the same type as ASP Logic Program rules. For example, translate 'apples are clean' as '\textlangle state\textrangle(clean(X)) :- object(X, apple).', where 'apple' is an object type and 'clean' is a predicate of arity 1 and argument type apple.
\\
\\
Rule 2. Represent properties of named objects as ASP Logic Program facts. For example,
translate sentence 'apple1 is red' as '\textlangle state\textrangle(red(apple1)).', where 'apple1' is an object and 'red' is a predicate of arity 1 and argument type apple.
\\
\\
Rule 3. Represent a relationship all unnamed objects of an object type with a named object as a ASP Logic Program rule. For example, translate 'apples are on the table' as '\textlangle state\textrangle(on(X, table)) :- object(X, apple).' where 'apple' is an object type, 'table' is an object of type 'furniture' and 'on' is a predicate of arity 2 and argument types ['apple', 'furniture'].
\\
\\
Rule 4. Represent relationships between two or more named objects as ASP Logic Program facts. For example, translate sentence 'apple1 is on plate2' as '\textlangle state\textrangle(on(apple1, plate2)).', where 'apple1' and  'plate2' are objects of type 'apple' and 'plate' resp, and 'on' is a predicate of arity 2 and argument types ['fruit, 'plate'].
\\
\\
Rule 5. Represent relationships between two or more unnamed objects as ASP Logic Program rules. For example, translate sentence 'cups handles are free' as '\textlangle state\textrangle(free(X, Y)) :- cup\_handle(X, Y).', where 'cup\_handle' is a predicate of arity 2 and argument types ['cup', 'handle'] and 'free' is a predicate of arity 2 and argument types ['cup', 'handle'].
\\
\\
Rule 6. Represent a bijective map between two sets of unnamed objects as ASP Logic Program facts.
For example, translate sentence '3 balls for 3 kids' as '\textlangle state\textrangle(map(ball, play, kid)).', where 'ball' and 'kid' are object types, and 'play' is a predicate of arity 2 with argument types ['ball', 'kid']. For example, translate sentence 'each cell has a ball' as 
'\textlangle state\textrangle(map(cell, cell\_ball\_map, ball)).', where 'cell' and 'ball' are object types, and 'cell\_ball\_map' is a bijective map predicate of arity 2 with argument types ['cell', 'ball'].
\\
\\
Rule 7. Represent objects that denote a cell on a two dimensional grid 'cell\_grid' as
ASP Logic Program facts. For example, translate objects 'cell\_2\_3' as '\textlangle state\textrangle(cell\_grid(2, 3, cell\_2\_3)).', where 'cell\_2\_3' is an object and 'cell\_grid' is a provided tertiary predicate.
\end{tcolorbox}

\begin{tcolorbox}[enhanced, breakable, title=LLM Prompt for Rules Extraction Ctd.,size=fbox]
Below an example of objects, predicates, text and the answer.
\\
\\
Objects:
\begin{verbatim}
{'apple': ['apple1', 'apple2'], 'orange': [], 'plate': 
['plate_1', 'plate_2', 'plate_3', 'plate_4', 'plate_5'],
'table': ['table1', 'table2'], 'ball': ['ball1', 
'ball3'], 'room': ['hall'], 'kid': [], 'robot': [], 
'cell': ['cell_0_1', 'cell_0_2', 'cell_1_1', 'cell_1_2', 
'cell_2_1', 'cell_2_2', 'cell_3_1', 'cell_3_2']}
\end{verbatim}
Predicates: [AS ABOVE]
\\
\\
Text: [AS ABOVE]
\\
\\
Answer:
\begin{verbatim}
init(on(apple1, plate_2)).
init(on(apple2, plate_1)).
init(on(orange1, plate_1)).
init(on(orange3, plate_2)).
init(in_1(ball1, table1)).
init(in_2(ball3, table1)).
init(in_1(apple3, table2)).
init(in_2(apple2, table2)).
init(red(X)) :- object(X, apple).
init(map(ball, play, kid).
init(clean(X)) :- object(X, plate).
init(inside(X, hall)) :- object(X, kid).
init(cell_grid(0, 1, cell_0_1)).
init(cell_grid(0, 2, cell_0_2)).
init(cell_grid(1, 1, cell_1_1)).
init(cell_grid(1, 2, cell_1_2)).
init(cell_grid(2, 1, cell_2_1)).
init(cell_grid(2, 2, cell_2_2)).
init(cell_grid(3, 1, cell_3_1)).
init(cell_grid(3, 2, cell_3_2)).
init(map(cell, cell_robot_map, robot)).
goal(on(orange1, plate_3)).
goal(on(orange2, plate_4)).
goal(on(orange3, plate_5)).
goal(not_clean(X)) :- object(X, plate).
goal(map(kid, paint, ball)).
goal(inside(X, hall)) :- object(X, kid).
goal(not unplated)).


\end{verbatim}
Objects: [AS EXTRACTED IN THE OBJECT EXTRACTION STEP]
\\
\\
Predicates: \textlangle Predicates of the task domain\textrangle
\\
\\
Text: \textlangle Task description\textrangle
\\
\\
Answer:
\end{tcolorbox}

\begin{tcolorbox}[enhanced, breakable, title=Generic LLM Prompt for generating intermediate representation in one step,size=fbox]
Given
\\
    (1) A JSON list of object types with their descriptions.
\\
    (2) A list of predicates. The predicates of arity one represent properties of objects.
    The predicates of arity two or more represent relationships of two or more objects.
\\
    (3) A text describing a planning task. A planning task description consists of initial
    state and goal state descriptions.
\\
\\
Translate the provided text following below steps.
\\
Step 1. Detect the initial state and goal state descriptions. 
\\
\\
Step 2. Extract for each object type T the explicitly mentioned count, width, and height of instances of T from the init state description. 
\\
\\
Step 3. Extract all instances of each object type from the provided text. Instances appear as singular nouns in the text. Ignore the plural nouns. In order to do so, consider 
        only the instances that are explicitly mentioned in the provided text. Do not output instances that 
        are not mentioned in the text. If an instance contains hyphens, replace the hyphens by underscores. If an instance
        begins with an upper case, then lower case it. If an instance contains spaces, replace the spaces by underscores.
\\
\\
Step 4. Translate each state description by following below rules as applicable.
        '\textlangle state\textrangle' is a placeholder for 'init' or 'goal' for sentences of initial state or 
        goal state respectively. While doing so, use only the provided predicates and respect their arity and type of arguments. Also use the object types and list of objects extracted in Step 3.
\\
\\
Step 5. Output the result as a valid ASP Program without any comments or additional text.
\\
\\
State sentence translation rules: [SAME AS ABOVE]
\\
\\
Object types: [SAME AS ABOVE]
\\
\\
Predicates: [SAME AS ABOVE]
\\
\\
Text: [SAME AS ABOVE]
\\
\\
Answer: [SAME AS ABOVE]
\\
\\
Objects types: \textlangle Object types of the task domain\textrangle
\\
\\
Predicates: \textlangle Predicates of the task domain\textrangle
\\
\\
Answer:
\end{tcolorbox}

\section{PDDL Comparision}
\label{appendix:comparison}
Following algorithm illustrate the main steps of for comparing two PDDLs. We use this to compare the TIC generated task PDDL with the ground truth task PDDL.

\begin{algorithm}[h!]
    \centering
    \begin{algorithmic}[1]
    \State taskPDDLTypes $\gets$ getPDDLTypes(taskPDDL)
    \State correctPDDLTypes $\gets$ getPDDLTypes(correctPDDL)
    \If{set(taskPDDLTypes) $\neq$ set(correctPDDLTypes)}
        \State \Return $False$
    \EndIf
    \State taskPDDLObjects $\gets$ getPDDLObjects(taskPDDL)
    \State correctPDDLObjects $\gets$ getPDDLObjects(correctPDDL)
    \If{len(taskPDDLObjects) $\neq$ len(correctPDDLObjects)}
        \State \Return $False$
    \EndIf
    \State taskInitStateAtoms $\gets$ getInitStateAtoms(taskPDDL)
    \State correctInitStateAtoms $\gets$ getInitStateAtoms(correctPDDL)
    \If{len(taskInitStateAtoms) $\neq$ len(correctInitStateAtoms)}
        \State \Return $False$
    \EndIf
    \State taskGoaltStateAtoms $\gets$ getGoalStateAtoms(taskPDDL)
    \State correctGoalStateAtoms $\gets$ getGoalStateAtoms(correctPDDL)
    \If{len(taskGoalStateAtoms) $\neq$ len(correctGoalStateAtoms)}
        \State \Return $False$
    \EndIf
    \State taskObjects $\gets$ []
    \State correctObjects $\gets$ []
        
    \ForAll{t $\in$ sort(taskPDDLTypes)}
        \State correctObjectsOfType $\gets$ getObjectsOfType(correctPDDL, t)
        \State typePerm $\gets$ permutations(correctObjectsOfType) 
        \State append typePerm to correctObjects
        \State taskObjectsOfType $\gets$ getObjectsOfType(taskPDDL, t)
        \State append taskObjectsOfType to taskObjects
    \EndFor

    \ForAll{perm $\in$ cartesianProduct(correctObjects)} \hfill{//each perm is a list containing one permutation of objects for each object type} 
        \If{checkStateEquivalence(taskObjects, perm, taskInitStateAtoms, correctInitStateAtoms) and\\
        checkStateEquivalence(taskObjects, perm, taskGoaltStateAtoms, correctGoalStateAtoms}
            \State \Return $True$      
        \EndIf
    \EndFor
    \State \Return $False$
    \end{algorithmic}
    \caption{Compare PDDLs}
    \label{alg:fixlabelsmain}
\end{algorithm}

checkStateEquivalence of two states $s$ and $t$ is true if state $s$ can be simulated by state $t$ and vice versa, otherwise false. A state $s$ is simulated by another state $t$ if each atom in $s$ can be simulated by an atom in state $t$. That is, there exists a mapping $m$ from objects of state $s$ to objects of state $t$ such that for every atom $a \in s$ there exists an atom $b \in t$ such that $m(a) = b$.

\section{Domain Knowledge as ASP Rules}
\label{sec:domain-rules}

% \begin{scriptsize}
    
\subsection{Barman}

\begin{verbatim}
init(@make_seq(N-1, l, next, 1)) :- cardinality(level, N).
first_level(X) :- object(X, level), not init(next(_, X)).

init(shaker_empty_level(X, Y)) :- object(X, shaker), 
 first_level(Y).
init(shaker_level(X, Y)) :- object(X, shaker), first_level(Y).

object(X, hand) :- init(handempty(X)).

object(L, level) :- init(next(L, _)).
object(L, level) :- init(next(_, L)).

object(X, cocktail) :- init(cocktail_part1(X, _)).
object(X, cocktail) :- init(cocktail_part2(X, _)).
object(X, ingredient) :- init(cocktail_part1(_, X)).
object(X, ingredient) :- init(cocktail_part2(_, X)).

object(X, shot) :- goal(contains(X, _)).    
\end{verbatim}

\subsection{Blocksworld}
\begin{verbatim}
object(B, block) :- init(on(B, _)).
object(B, block) :- init(on(_, B)).
object(B, block) :- init(on_table(B)).
object(B, block) :- init(clear(B)).

object(B, block) :- goal(on(B, _)).
object(B, block) :- goal(on(_, B)).
object(B, block) :- goal(on_table(B)).
object(B, block) :- goal(clear(B)).

object(B, object) :- object(B, block).
\end{verbatim}

\subsection{Floortile}
\begin{verbatim}
floortile_grid(R,C,Z1) :- init(floortile_grid(R,C,Z1)).
init(up(Z1,Z2)) :- floortile_grid(R,C,Z1), 
 floortile_grid(R-1,C,Z2).
init(down(Z2,Z1)) :- floortile_grid(R,C,Z1), 
 floortile_grid(R-1,C,Z2).
init(right(Z1,Z2)) :- floortile_grid(R,C,Z1), 
 floortile_grid(R,C-1,Z2).
init(left(Z2,Z1)) :- floortile_grid(R,C,Z1), 
 floortile_grid(R,C-1,Z2).
init(robot_has(robot1, white)).
init(robot_has(robot2, black)).

object(Z, tile) :- floortile_grid(_,_,Z).

object(X, color) :- init(robot_has(_, X)).
object(X, robot) :- init(robot_has(X, _)).

object(X, color) :- init(available_color(X)).
object(X, robot) :- init(robot_at(X, _)).
object(X, tile) :- init(robot_at(_, X)).

object(X, color) :- goal(painted(_, X)) 
object(X, tile) :- goal(painted(X, _)).

init(clear(T)) :- object(T, tile), not init(robot_at(_, T)).

init(min_cost_metric("=(total-cost) 0)")).

init(available_color(X)) :- init(robot_has(_, X)).
init(available_color(X)) :- goal(painted(_, X)).
\end{verbatim}

\subsection{Grippers}
\begin{verbatim}
object(X, robot) :- init(at_robby(X, _)).
object(X, room) :- init(at_robby(_, X)).
object(X, ball) :- init(at(X, _)).
object(X, room) :- init(at(_, X)).
object(X, ball) :- goal(at(X, _)).
object(X, room) :- goal(at(_, X)).
object(X, object) :- object(X, ball).
object(X, gripper):- init(free(_, X)).
object(X, robot):- init(free(X, _)).
object(X, gripper) :- object(X, left_gripper).
object(X, gripper) :- object(X, right_gripper).

robot_left_gripper_map(X,Y) :- init(robot_left_gripper_map(X,Y)).
robot_right_gripper_map(X,Y) :- 
 init(robot_right_gripper_map(X,Y)).
\end{verbatim}

\subsection{Storage}
\begin{verbatim}
object(X, storearea) :- init(on(_, X)).
object(X, crate) :- init(on(X, _)).
object(X, hoist) :- init(at(X, _)).
object(X, storearea) :- object(X, depot_storearea).
depot_storearea_grid(R,C,Z) :- init(depot_storearea_grid(R,C,Z)).

adjacent(Z1,Z2) :- depot_storearea_grid(R,C,Z1), 
 depot_storearea_grid(R-1,C,Z2).
adjacent(Z2,Z1) :- depot_storearea_grid(R,C,Z1), 
 depot_storearea_grid(R-1,C,Z2).
adjacent(Z1,Z2) :- depot_storearea_grid(R,C,Z1), 
 depot_storearea_grid(R,C-1,Z2).
adjacent(Z2,Z1) :- depot_storearea_grid(R,C,Z1), 
 depot_storearea_grid(R,C-1,Z2).

init(connected(X, Y)) :- init(connected(Y, X)).

goal(connected(X, Y)) :- goal(connected(Y, X)).
\end{verbatim}

\subsection{Termes}
\begin{verbatim}
init(@make_seq(N, n, succ, 0)) :- init(max_height(N)).
object(N, numb) :- init(succ(N, _)).
object(N, numb) :- init(succ(_, N)).
termes_pos_grid(R,C,Z) :- init(termes_pos_grid(R,C,Z)).

object(Z, position) :- termes_pos_grid(_,_,Z).
init(height(Z, n0)) :- termes_pos_grid(_,_,Z).

init(neighbor(Z1,Z2)) :- termes_pos_grid(R,C,Z1), 
 termes_pos_grid(R-1,C,Z2).
init(neighbor(Z2,Z1)) :- termes_pos_grid(R,C,Z1), 
 termes_pos_grid(R-1,C,Z2).
init(neighbor(Z1,Z2)) :- termes_pos_grid(R,C,Z1), 
 termes_pos_grid(R,C-1,Z2).
init(neighbor(Z2,Z1)) :- termes_pos_grid(R,C,Z1), 
 termes_pos_grid(R,C-1,Z2).

0 { goal(height(Z, n0)) } 1 :- termes_pos_grid(_,_,Z).
:- termes_pos_grid(_,_,Z), #count {X : goal(height(Z, X))} != 1.
\end{verbatim}

\subsection{Tyreworld}
\begin{verbatim}
init(@make_map(T1, P, T2, N)) :- init(map(T1, P, T2)), 
 cardinality(T1, N).
object(@gen_objects(N, 0, intact_tyre), intact_tyre) :- 
 cardinality(intact_tyre, N).
object(@gen_objects(N, 0, flat_tyre), flat_tyre) :- 
 cardinality(flat_tyre, N).
object(@gen_objects(N, 0, hub), hub) :- cardinality(hub, N).
object(@gen_objects(N, 0, nut), nut) :- cardinality(nut, N).
goal(@make_map(T1, P, T2, N)) :- goal(map(T1, P, T2)), 
 cardinality(T1, N).

object(X, container) :- init(in(_, X)).
object(X, wheel) :- object(X, flat_tyre).
object(X, wheel) :- object(X, intact_tyre).
object(X, tool):- object(X, jack).
object(X, tool):- object(X, pump).
object(X, tool):- object(X, wrench).
init(intact(X)) :- object(X, intact_tyre).

object(X, container) :- init(in(_, X)).

object(X, hub) :- init(on(_, X)).

object(X, nut) :- init(tight(X,_)).
object(X, hub) :- init(tight(_,X)).
\end{verbatim}
% \end{scriptsize}

\section{Examples of Language Variation}
\label{appendix: language variation example}
Below we provide a few examples of Language Variation of task descriptions. Please refer to supplementary material for the entire Language Variation dataset.

\subsection{}
\begin{tcolorbox}[enhanced, breakable,title=Barman task description from original dataset ,size=fbox]
You have 1 shaker with 3 levels, 4 shot glasses, 3 dispensers for 3 ingredients. 
The shaker and shot glasses are clean, empty, and on the table. Your left and right hands are empty. 
The first ingredient of cocktail1 is ingredient2. The second ingredient of cocktail1 is ingredient1. 
The first ingredient of cocktail2 is ingredient2. The second ingredient of cocktail2 is ingredient3. 
The first ingredient of cocktail3 is ingredient1. The second ingredient of cocktail3 is ingredient2. 
Your goal is to make 3 cocktails. 
shot1 contains cocktail1. shot2 contains cocktail3. shot3 contains cocktail2. 
\end{tcolorbox}

\begin{tcolorbox}[enhanced, breakable,title=Barman task description from Language Variation dataset ,size=fbox]
There is a shaker with 3 levels, 4 shot glasses, and 3 dispensers for 3 ingredients.
The shaker and the shot glasses are on the table.
The shaker and the shot glasses are clean and empty. Your left and right hands are empty.
There are 3 cocktails to be made, each with 2 ingredients.
The first cocktail requires ingredient2 as the first ingredient and ingredient1 as the second ingredient.
The second cocktail requires ingredient2 as the first ingredient and ingredient3 as the second ingredient.
The third cocktail requires ingredient1 as the first ingredient and ingredient2 as the second ingredient.
Your task is to make the 3 cocktails and pour them into shot glasses. The first cocktail should be poured into shot glass 1, the second cocktail into shot glass 3, and the third cocktail into shot glass 2.

\end{tcolorbox}

\subsection{}
\begin{tcolorbox}[enhanced, breakable,title=Floortile task description from original dataset, size=fbox]
You have 5 rows and 3 columns of unpainted floor tiles. 
\\
tile-0-1 tile-0-2 tile-0-3 
\\
tile-1-1 tile-1-2 tile-1-3 
\\
tile-2-1 tile-2-2 tile-2-3 
\\
tile-3-1 tile-3-2 tile-3-3 
\\
tile-4-1 tile-4-2 tile-4-3 
\\
You have 2 robots. 
\\
Each robot can paint in color white or black.
\\
robot2 is at tile-1-1. 
\\
robot1 is at tile-2-3. 
\\
Your goal is to paint the grid in the following pattern: 
tile-1-1 is white; tile-1-2 is black; tile-1-3 is white; tile-2-1 is black; tile-2-2 is white; tile-2-3 is black; tile-3-1 is white; tile-3-2 is black; tile-3-3 is white; tile-4-1 is black; tile-4-2 is white; tile-4-3 is black. 
\end{tcolorbox}

\begin{tcolorbox}[enhanced, breakable,title=Floortile task description from Language Variation dataset, size=fbox]
There are 15 unpainted floor tiles arranged in 5 rows and 3 columns.
\\
tile-0-1 tile-0-2 tile-0-3
\\
tile-1-1 tile-1-2 tile-1-3
\\
tile-2-1 tile-2-2 tile-2-3
\\
tile-3-1 tile-3-2 tile-3-3
\\
tile-4-1 tile-4-2 tile-4-3
\\
Two robots are available to paint the tiles either white or black.
\\
Robot 2 is currently positioned at tile-1-1 while robot 1 is at tile-2-3.
\\
The objective is to paint the tiles in a specific pattern where tile-1-1 is white, tile-1-2 is black, tile-1-3 is white, tile-2-1 is black, tile-2-2 is white, tile-2-3 is black, tile-3-1 is white, tile-3-2 is black, tile-3-3 is white, tile-4-1 is black, tile-4-2 is white, and tile-4-3 is black.
\end{tcolorbox}

\subsection{}
\begin{tcolorbox}[title=Grippers task description from original dataset, size=fbox]
You control 3 robots, each robot has a left gripper and a right gripper. 
\\
There are 4 rooms and 7 balls. 
\\
robot3 is in room3. 
\\
robot2 is in room1. 
\\
robot1 is in room4. 
\\
ball1 is in room4. ball7 is in room3. ball3 is in room3. ball4 is in room2. ball6 is in room1. ball2 is in room3. ball5 is in room2. 
\\
The robots' grippers are free. 
\\
Your goal is to transport the balls to their destinations. 
\\
ball1 should be in room1. 
\\
ball2 should be in room4. 
\\
ball3 should be in room3. 
\\
ball4 should be in room4. 
\\
ball5 should be in room3. 
\\
ball6 should be in room2. 
\\
ball7 should be in room1. 
\end{tcolorbox}

\begin{tcolorbox}[enhanced, breakable,title=Grippers task description from Language Variation dataset, size=fbox]
There are 3 robots. Each robot has a left gripper and a right gripper.
\\
 There are 4 rooms and 7 balls.
 \\
 Robot 3 is in room 3, robot 2 is in room 1, and robot 1 is in room 4.
 \\
 The balls are distributed among the rooms as follows: ball 1 is in room 4, ball 7 is in room 3, ball 3 is in room 3, ball 4 is in room 2, ball 6 is in room 1, ball 2 is in room 3, and ball 5 are in room 2.
 \\
 The robots' grippers are currently not holding anything.
 \\
 The objective is to move the balls to their designated rooms: ball 1 to room 1, ball 2 to room 4, ball 3 to room 3, ball 4 to room 4, ball 5 to room 3, ball 6 to room 2, and ball 7 to room 1.

\end{tcolorbox}

%\end{subappendices}

\end{document}